\RequirePackage{fix-cm}
\documentclass{svjour3}                     
\smartqed  
\usepackage{graphicx}
\usepackage{times}
\usepackage{helvet}
\usepackage{courier}
\usepackage{subcaption}
\usepackage[linesnumbered,ruled,vlined]{algorithm2e}
\usepackage{amsmath,amssymb,amsfonts}
\usepackage[usenames, dvipsnames, table]{xcolor}

\usepackage[markup=bfit, deletedmarkup=sout, authormarkup=superscript]{changes}
\usepackage[hang,flushmargin]{footmisc}
\definechangesauthor[name={at}, color= Green] {ADD}
\definechangesauthor[name={bh}, color=purple] {BH}
\definechangesauthor[name={del}, color= red] {DEL}
\definechangesauthor[name={td}, color= Sepia] {TODO}
\newcommand{\at}[1]{\added[id=  ADD]{#1}}

\begin{document}
\title{One-shot Policy Elicitation via Semantic \\ Reward Manipulation
}


\author{Aaquib Tabrez         \and
        Ryan Leonard         \and
        Bradley Hayes
}


\institute{
\at
Department of Computer Science, University of Colorado Boulder, Boulder, CO 80309, USA \and
Aaquib Tabrez (Corresponding Author)\at
\email{mohd.tabrez@colorado.edu}
\and
Ryan Leonard \at
\email{ryan.leonard@colorado.edu}
\and
Bradley Hayes \at
\email{bradley.hayes@colorado.edu}
}

\date{Received: date / Accepted: date}

\maketitle

\begin{abstract}
Synchronizing expectations and knowledge about the state of the world is an essential capability for effective collaboration. For robots to effectively collaborate with humans and other autonomous agents, it is critical that they be able to generate intelligible explanations to reconcile differences between their understanding of the world and that of their collaborators. In this work we present Single-shot Policy Explanation for Augmenting Rewards (SPEAR), a novel sequential optimization algorithm that uses semantic explanations derived from combinations of planning predicates to augment agents' reward functions, driving their policies to exhibit more optimal behavior. We provide an experimental validation of our algorithm's policy manipulation capabilities in two practically grounded applications and conclude with a performance analysis of SPEAR on domains of increasingly complex state space and predicate counts. We demonstrate that our method makes substantial improvements over the state-of-the-art in terms of runtime and addressable problem size, enabling an agent to leverage its own expertise to communicate actionable information to improve another's performance. 
\end{abstract}

\section{Introduction}

Autonomous systems have been shown to to be capable of motivating behavior changes and conveying new knowledge to improve human or agent performance on a multitude of tasks \cite{belpaeme2018social,leite2013social,leyzberg2014personalizing}. 
Unfortunately, the process of generating concise and informative explanations capable of eliciting desired changes is a difficult task, as it requires both insights into a collaborator’s decision making process and the ability to determine and convey important information. Further, repairing a human policy requires the generation of explanations which are intelligible and concise, especially in situations with high costs of failure or time pressure \cite{hayes2017hri,chakraborti2019plan,tabrez2020automated}. Similarly, for autonomous agents operating with different state representations, policy repair requires a common ground (language) to communicate updates for efficient behavior modification during the task.

\begin{figure}[t]%
	\centering%
	\begin{subfigure}[h]{\textwidth}
        \includegraphics[width=\textwidth]{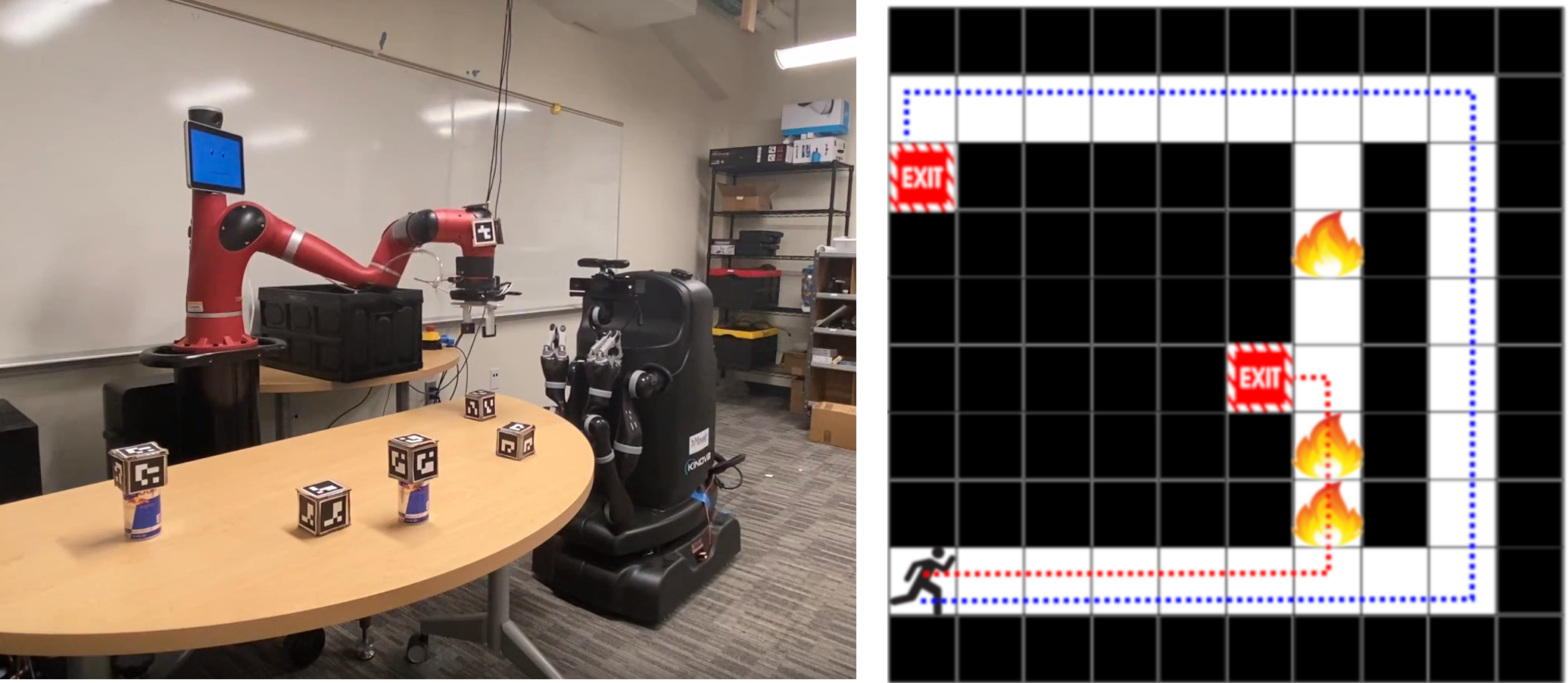}
    \end{subfigure}\\%
    \centering%
    \caption{ (Left) The red robot is engaged in a tabletop cleaning task but has a malformed policy that will throw away all of the objects, despite some not being garbage. (Right) A human agent attempts to navigate to a building exit in an emergency evacuation scenario lacking knowledge about the location of the fires. Using SPEAR, these policies can be repaired, leveraging semantic updates to produce more optimal behavior. 
    }
    \label{true-map}
\end{figure}

In this work we provide a characterization of the problem of semantically manipulating an agent's policy through external advice, and propose a solution that provides autonomous agents with the ability to: 1) Infer the reward function that most likely drives another agent's behavior; 2) Identify divergences between another agent's reward function and their own; and 3) Communicate advice to repair relevant differences concisely and efficiently at an appropriate level of detail, removing harmful behaviors from the other agent's policy.

An illustrative scenario that we use to showcase the importance and practicality of this work is routing people during an emergency evacuation, particularly in cases where habitants of the building are aware of neither the precise nature of the emergency nor the entire building's floorplan.
Even if people are capable of navigating towards the nearest exit, uncertainty about the nature of the environment and its hazards could be disasterous. Adding to the complexity of time-critical, high-consequence scenarios like this one, it is essential that any advice or instructions account for the recipient's knowledge of the world, and therefore must also consider tradeoffs between accuracy, specificity, and interpretability \cite{robinette2016overtrust,lage2019evaluation}. As an example, someone visiting a building for a meeting may not know how to change their evacuation plan when told ``There's a fire near Conference Room 3'', but may be able to adapt their plan if told ``the north half of the building is on fire'' for the same scenario. Even though the latter phrase may not communicate the true representation of the hazard, it is more easily comprehensible to someone who is less familiar with the building than the former. Thus, autonomous systems that aim to generate useful feedback to help others will need to explicitly consider the complexity of their own explanations and the knowledge held by those they attempt to help.

To achieve this, we present Single-shot Policy Explanation for Augmenting Rewards (SPEAR), a novel Integer Programming-based algorithm for generating reward updates in the form of semantic advice to improve the policies of human and robot collaborators. SPEAR enables an autonomous robot to utilize knowledge about the beliefs and goals of its collaborators to identify inaccuracies in their models, generating targeted, human-interpretable guidance for updating their reward functions (and thus, policies) during a task. Key to its effectiveness, SPEAR generates feedback with levels of specificity appropriate to the agent's understanding of the world. The four primary contributions of our work are:

\begin{itemize}

\item A characterization of the policy elicitation problem.

\item SPEAR, a novel algorithm for improving task performance through semantic elicitation of others' policies. 

\item A linear programming formulation enabling semantic communication of world state regions, achieving a several order of magnitude improvement over the nearest comparable state-of-the-art \cite{hayes2017hri} in terms of computation time and applicable domain size.

\item An experimental validation and performance analysis of SPEAR, grounded in practical application scenarios.

\end{itemize}

\renewcommand{\thefootnote}{\fnsymbol{footnote}}
\section{Background and Related Work}

As autonomous systems become increasingly capable decision makers, Explainable AI (xAI) has emerged as a necessary component for fielding safe autonomous systems. Many of today's systems with learned control policies continue to only satisfy Donald Michie's weak criterion for learning systems\footnote[1]{
	In 1980, Donald Michie proposed three criteria to evaluate machine learning systems \cite{michie1988machine}:\\
	1. \underline{Weak Criterion}: A system increases performance on unseen data by learning from sample data.\\
	2. \underline{Strong Criterion}: The system contains the weak criterion plus the ability to communicate its learned hypotheses function in symbolic form.\\
	3. \underline{Ultra-strong Criterion}: The system contains the strong criterion plus the ability to teach a user the learned hypothesis function.
	},
despite explainability being recognized as crucial for improving transparency, trust, and team performance \cite{vigano2018explainable,miller2019explanation,wang2016trust}, further underscoring the need for the development of xAI techniques that work alongside popular, generally opaque methods. 


Furthermore, these black-box models are difficult to validate and debug when developers do not know what is happening inside the system, leading to further distrust and resistance towards the widespread usage of these algorithms \cite{arnold2019factsheets,sokol2020explainability}. Most of the research in explainable AI has been explicitly focused on making algorithms transparent to developers, allowing them the ability to debug or predict model behavior in restricted domains (e.g., how model parameters affect the final classification decision) \cite{ribeiro2016should,selvaraju2017grad}. These approaches are fundamentally limiting to non-expert stakeholders and/or end-users who interact with the models or products regularly, and thus directly experience the consequences of failures \cite{doshi2017towards,mittelstadt2019explaining}. Therefore, to generate explanations that are comprehensible and useful for both experts as well as non-experts, they should draw from characteristics of everyday human-centric explanations focused on the `why' of particular events, properties, or decisions, rather than general scientific relationships \cite{miller2019explanation,henne2019counterfactual}.

One popular approach in human-robot collaboration has been to use explanations to speed up the learning process or to explore alternative policies. These approaches primarily focus on active learning frameworks for modelling user preferences by asking questions \cite{racca2018active} or taking instruction \cite{fukuchi2017autonomous} in natural language. While these systems focus on improving a robotic agent's transparency and behavior using explanation, they don't account for collaborators' level of knowledge or need for the information. Therefore, these traditional goal-based agents lack the ability to consider a shared mental model in their planning, a crucial element for building effective teammates and interpretable behavior \cite{chakraborti2017ai,tabrez2020survey}. 

Another trend in the direction of ``Value of Information" (VOI) in human-robot interaction has been to decide what information should be communicated and when it should be delivered in the collaborative decision making process \cite{kaupp2010human}. Kaupp et al. used VOI theory to propose a framework in which robots query human operators in uncertain environments if the expected benefit of the humans' feedback exceeds the cost of the query. They show that accounting for human cost in querying has an advantage in performance, operator workload, usability, and the users’ perception of the robot. More recently, active learning has been utilized to understand user preference and provide valuable information while keeping trust and adaptability of humans in account \cite{chitnis2018learning,racca2018active,dorsa2017active}.

Identifying and resolving the model differences of a collaborator (i.e., establishing a shared mental model) is a key aspect of establishing interpretability \cite{hayes2015iros,chakraborti2017ai,dragan2017robot}. Having interpretable and predictable behavior is essential for achieving team fluency and avoiding catastrophic failures in planning and decision making \cite{hayes2013collab,nikolaidis2013human}.

An effective approach for establishing shared mental models in human-robot collaboration has been to use natural language to explain robots' behavior or underlying logic \cite{hayes2017hri,tabrez2019explanation}. In \cite{hayes2017hri}, the authors formulated the problem of efficiently describing state regions as a set cover problem, trying to find the smallest logical expression of predicates that succinctly describe a target state region. Their approach is innovative in performing explanation generation using communicable predicates, but their solution is exponential in memory and runtime with respect to the size of the domain and predicate set, preventing its use in most real-world problems. 

Others have explored the generation of different types of explanations based on user preference \cite{zahedi2019towards}. Work by Briggs and Scheutz explored adverbial cues informed by Grice’s maxims \cite{grice1975logic} of effective conversational communication (quality, quantity, and relation) to transparently track and update mental models of the collaborators \cite{briggs2011facilitating}. Van et al. developed transparent Reinforcement Learning (RL) models for the generation of contrastive explanations of agent behavior based on the user’s query-derived policy, and of the learned policy of the agent \cite{van2018contrastive}. They show that people are more interested in explanations about the behavior of an agent rather than a single action taken by it. People have also been shown to prefer contrastive, selective, and social properties in explanation, as previously hinted in the psychology literature \cite{chakraborti2019plan,ciocirlan2019human,miller2019explanation}. 

In this work we focus on \emph{policy elicitation}, a process through which feedback is crafted and given to another agent, in the form of reward function updates during task execution, such that they change their behavior to match the desired policy. By providing reward information for targeted regions of state space through explanations (symbolic updates), we can modify a collaborator's reward function using semantic descriptions. This allows our method to operate at a level of abstraction that does not depend on the recipient's underlying state space representation, instead only requiring a similar vocabulary of planning predicates.

\begin{figure*}[b]%
	\centering%
	\includegraphics[width=\textwidth]{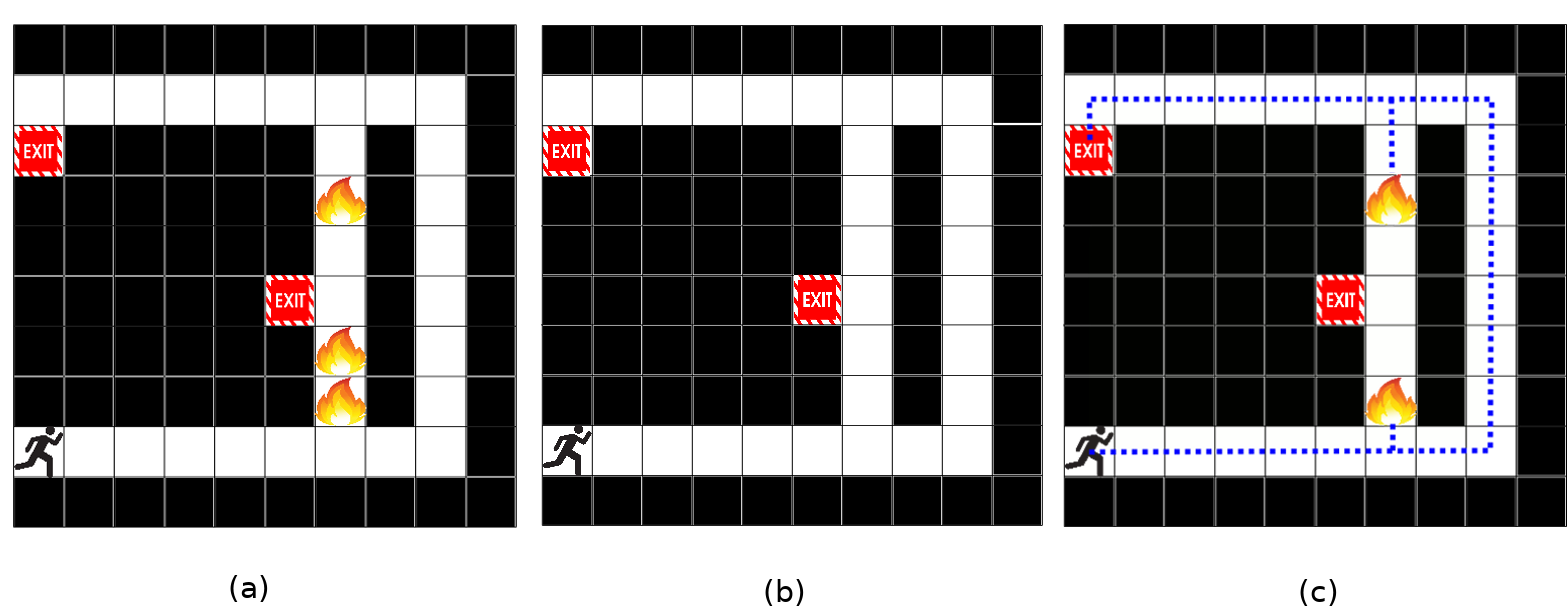}

	\caption{Rollouts of the agent's estimated policy are used to identify problematic behavior. (a) The true environment that the agent is acting in. (b) The environment that the agent believes it is acting in. (c) A rollout of the agent's estimated policy reveals the minimal amount of hazardous states that need to be communicated for policy repair.
}
\label{fig:policy-elicitation}
\end{figure*}

\begin{figure*}[h]%
	\centering%
	\includegraphics[width=\textwidth]{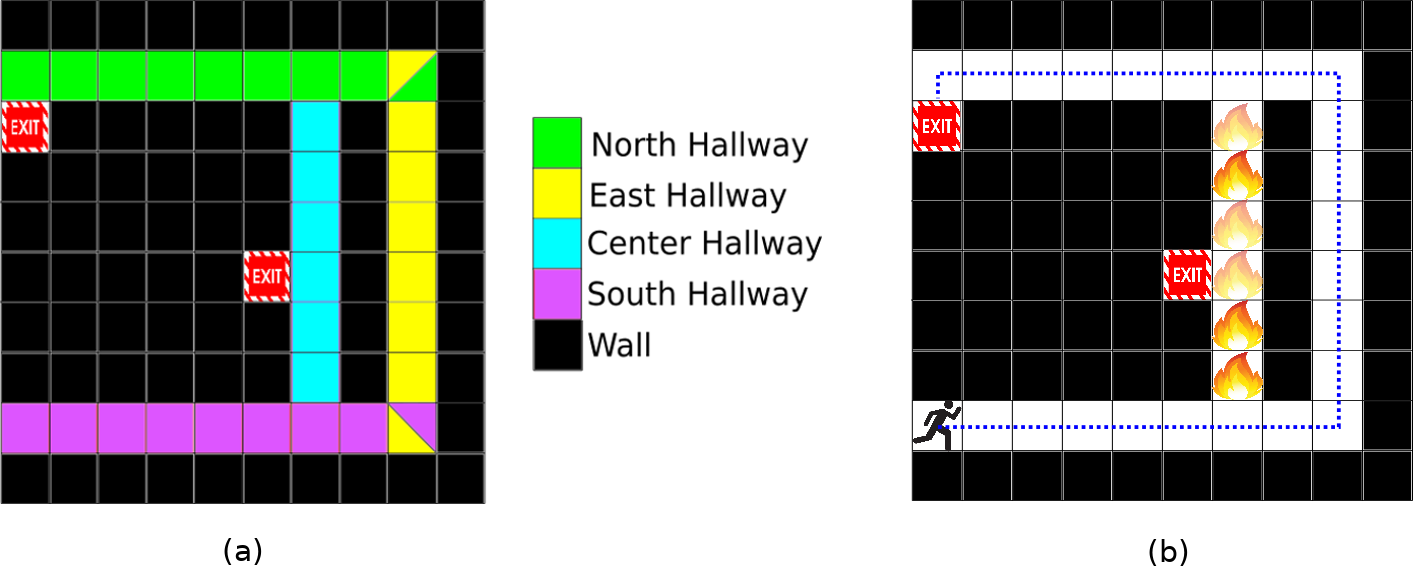}

	\caption{A Boolean algebra over predicates with string descriptions is used as the basis for grounding reward function updates in language. (a) A map of the environment with an overlay indicating states where various predicates are true. (b) A potential belief update for the human's reward function, realized after communicating ``The center hallway is on fire". Some of the fires, indicated by being faded, don't actually exist in the environment and are an artifact of imprecise language. Despite this imprecision, the optimal policy can be elicited from this update.}
	
   
\label{fig:policy-elicitation2}
\end{figure*}

\section{Policy Elicitation via Social Manipulation} \label{policy-elicitation}

The goal of policy elicitation is to cause a behavior change (policy update) in another agent through some form of communicative act. To effectively collaborate with others and coach them towards more optimal policies, it is essential that these communications are intelligible \cite{andre2002state,abel2018state,tabrez2019hripioneers}, but directly communicating states (i.e. the feature vector itself) relies on both the feature space between agents being the same and there being an efficient mechanism to communicate that information quickly. These criteria are unlikely to hold with either humans or heterogeneous autonomous agents (not all agents will have the same state representations) as the intended recipients. Our work generates state space-agnostic natural language descriptions of state regions and corresponding reward information, allowing agents to update their reward functions (and policies).

We define a \textbf{base predicate} to be a pre-defined boolean state classifier (as found in traditional STRIPS planning \cite{fikes1971strips}) with associated string explanation (e.g., in\_central\_hallway(x) $\rightarrow$ ``X is in the central hallway.''). To represent intersections of predicates (e.g., ``in the central hallway'' AND ``has fire extinguisher''), we introduce \textbf{composite predicates}, which consist of multiple base predicates and evaluate to true if and only if all base predicate members evaluate to true. Predicates may also have a cost associated with them (e.g., how long they take to communicate) to assist the optimizer with generating more desirable solutions. In SPEAR, predicates are used to communicate reward updates to improve collaborator task performance (Figure \ref{fig:policy-elicitation2}).

We characterize the problem of repairing or otherwise manipulating an agent's policy as one of identifying and reconciling divergence between a source and target reward function. Our proposed algorithm relies upon the following assumptions: 1) agents can understand the natural language description of the predicates; 2) agents are operating using a rational planner informed by a reward function (or something analogous); and 3) agents' suboptimal action selections are attributable to a malformed reward function rather than a malformed policy search algorithm or dynamics model.

\section{Approach} \label{approach}

In  this  section,  we  specify  the  problem  of  reparative policy elicitation: repairing  or manipulating  an  agent’s  policy  by  communicating corrections for model divergences. Our approach to policy elicitation uses three components that: 1) estimate the agent's reward function; 2) determine important reward function disparities that prevent desirable behavior; and 3) determine which states to provide reward updates for and communicate a corrective explanation.

\subsection{Belief Estimation with Active Observation} \label{belief-estimation}
We formulate our domain as a Markov Decision Process (MDP)\cite{bellman1957markovian}, wherein an agent acts to maximize an expected reward. $M$ is a Markov Decision Process defined by the 4-tuple $(S,A,T,R)$ where $S$ is the set of states in the MDP, $A$ is the set available actions, $T$ is a stochastic transition function describing the model's action-based state transition dynamics, and $R$ is the reward function $R: S \times A \times S \rightarrow \mathbb{R}$. Intuitively, $M$ serves as a simulator for an agent in the task domain. 

Belief estimation (inferring an agent's plan) is performed by leveraging the assumption that the agent has a rational planner and any sub-optimal behavior can be attributed to a misinformed understanding of the task's reward function $R_h$, where $R_h$ is the malformed reward function the agent is using \cite{tabrez2019explanation}. We assume the agent's dynamics model is correct $(T_h = T)$. 

Inferring an agent's reward function $R_h$ (e.g., the agent's beliefs about the task w.r.t. the world) is non-trivial and an active field of research. Tabrez et al. \cite{tabrez2019explanation} provides an approach to this problem by solving for whether the agent knows about particular components of the reward function; therefore if one can maintain a belief over which rewards another agent is aware of, their reward function may be able to be inferred. 

Intuitively, estimating $R_h$ is achieved by observing agent behavior and the goal states they approach --- in the emergency evacuation domain, this corresponds to observing their coverage of the building and which exits they are heading toward. Once we obtain $R_h$, we can simulate the MDP $M$ using $R_h$ to get the agent's expected policy $\pi_h$.

\subsection{Finding Important Model Divergences}

Once we have our belief of the agent's possible reward function $R_h$, we identify divergences between the optimal policy $\pi^*$  and the policy of the agent $\pi_h$ that would cause a reduction in the agent's expected cumulative reward. We do this by comparing agent policies trained on $R$ and $R_h$, where $R$ is the true reward function of the domain and $R_h$ is the reward function used by the agent. We find a set of states $\bar{S}$ about which we communicate updated reward information to augment $R_h$, such that a more optimal policy (closer to the expected reward of $\pi^*$) is elicited (Figure \ref{fig:policy-elicitation2}b). 

\subsection{Communicating State Regions} \label{ip-formulation}

We approach the problem of efficiently describing state regions as a set cover problem, trying to find the smallest logical expression of communicable predicates to succinctly describe target states as in  Hayes and Shah \cite{hayes2017hri}. Unlike prior work, we solve for the minimum set cover of the targeted state region using an Integer Program. The inputs to our IP formulation include:

\begin{itemize}

\item A set of state indices $\bar{S}$ that correspond to states with expected reward function divergence that needs to be communicated about for policy repair.

$\bar{S} = \{s_1, s_2, \cdots, s_{|\bar{S}|}  \}$,

\item A set of communicable predicates $\bar{P} = \{p_1, p_2, \cdots p_{|\bar{P}|} \}$ requiring the following for a solution to exist:\\

    $\exists\ \bar{Q} \subseteq \bar{P}$ such that $s \in \bar{S}$ is covered by a predicate in $\bar{Q} \text{, } \forall \text{ }s \in \bar{S}$ --- for every state that needs to be covered ($\bar{S}$), there exists a non-empty subset of predicates that can cover it.
    \item  A set of costs $\bar{C} = \{c_1, c_2, \cdots, c_{|\bar{P}|} \}$, such that $c \neq 0 \text{ } \forall\ c \in \bar{C}$ --- every predicate has non-zero cost associated with using it to update the agent's reward function $R_h$.

    \item  The desired trajectory for the agent $\bar{O} = \{ o_1, o_2, \cdots, o_{|\bar{O}|}\}$, where $o_i$ describes the state achieved after taking the $i^{th}$ action following the optimal policy from the start state.

\end{itemize}

The cost for each predicate can be customized per task. This is an important characteristic, as many factors may influence the cost of a predicate. One such criteria for defining cost can be the length of the string describing the predicate. Such a criteria could generate more easily understood explanations by imposing penalties for being too verbose.


A solution to the policy elicitation problem consists of selecting predicates to communicate reward information about specific state regions such that a more optimal policy is produced within some $\epsilon$ bound of the optimal policy's expected reward, $|E_R(\pi | R) - E_R(\pi | R_h)| \leq \epsilon$.  To minimize this objective while satisfying all the constraints, we define the mathematical formulation of our IP, which we refer to as \textbf{SPEAR-IP}, below:

\begin{equation}\label{eq:objective}
min\ \sum_{j=1}^{|\bar{P}|} c_jx_j + L  \sum_{k=1}^{|\bar{O}|} \sum_{j=1}^{|\bar{P}|} v_{kj}x_j 
\end{equation}

subject to 

\textcolor{black}{
\begin{equation}\label{eq:constraint}
 \sum_{j=1}^{|\bar{P}|} u_{ij}x_j \geq 1 \quad \forall\ i \in [1, |\bar{S}|]
\end{equation}
}

where we define the \textcolor{black}{$|\bar{S}| \times |\bar{P}|$} matrix \textcolor{black}{$U$} by

\begin{equation}\label{eq:a_matrix}
\textcolor{black}{u_{ij}}= 
\begin{cases}
    1,& \text{if } s_i \text{ is covered by } p_j\\
    0,              & \text{otherwise}
\end{cases}
\end{equation}
and, the $|\bar{O}| \times |\bar{P}|$ matrix \textcolor{black}{$V$} by 
\begin{equation} \label{eq:b_matrix}
\textcolor{black}{v_{kj}}= 
\begin{cases}
    1,& \text{if } o_k \text{ is covered by } p_j\\
    0,              & \text{otherwise}
\end{cases}
\end{equation}
where $i$ and $k$ are the indices into sets $\bar{P}$ and $\bar{O}$ respectively. $x_j \in \{0,1\}$ is a binary variable with $x_j = 1 $ meaning predicate $p_j$ is included in the cover and $x_j = 0$ indicating exclusion from the set cover. Equation \ref{eq:objective} is the objective function which needs to be minimized for the desired set cover based on cost. The first term of Equation \ref{eq:objective} ensures that the net cost of the predicates selected is as low as possible.

Equation \ref{eq:objective}'s second term heavily penalizes the objective function if the chosen set cover overlaps with the desired path, in the case of communicating a negative reward (for positive reinforcement this can be reversed). $L$ is a large positive number, and is used to encapsulate near-optimal behavior in the objective function (providing a soft constraint). The second term can be used separately to provide a hard constraint for finding a solution which will definitely elicit the desired path, but this approach inherently restricts the ability of an integer program to find solutions which would provide a \emph{near} optimal policy update. This second term in the formulation leads to \textbf{three possible cases}: 1) Set cover solution with low objective value; 2) No solution for the set cover; and 3) Set cover solution with high objective value. 

Cases 1 and 2 are simple and describe the ability of SPEAR-IP to solve the set cover. Case 3 is interesting and provides the option of further exploration to find an alternate solution with the given set of predicates. We can use the set cover from Case 3 to determine which states in the cover overlapped with the desired path. Using these states as a reference, we can simulate an alternate desired policy by penalizing these states in the true reward function as well, effectively coming up with a contingency for not having precise enough language available. This process of penalizing states through $R$ is repeated until either Case 1 or 2 is achieved (i.e. an immediate solution with low objective value or no solution for given a set of predicates). 

Equation \ref{eq:constraint} provides a hard constraint for the inclusion of states from $\bar{S}$ in the set cover. Equation \ref{eq:a_matrix}  defines the elements of matrix $U$, which encapsulates cover constraints from $\bar{S}$ (inclusion of all states). Equation \ref{eq:b_matrix} defines the elements of matrix $V$ using the desired trajectory $\bar{O}$, encapsulating the requirements for eliciting the desired policy. 

\setlength{\textfloatsep}{2pt}
\begin{algorithm}[!ht]
	\caption{Single-shot Policy Explanation for Augmenting Rewards (SPEAR)}
	\label{algo-reward-repair}
	\KwIn{MDP $(S,A,T,R)$, Min. Reward Threshold $R_L$, Agent Reward Function $R_h$, Current state $s_c$, Num. Rollouts k} 
	\KwOut{Semantic Reward Correction}
	$\bar{S} \gets \emptyset$; $L \gets $large scalar value\;
	$R_h^* \gets R_h$; // Best possible agent reward function \\
	\For{\text{rollout in range(1 to k)}}{
	$r_c \gets 0$; // Cumulative reward \\
	$\pi_h \gets $policy trained on $R_h$\;
	// Find states to update for best possible $R_h^*$  \\
    $s \gets s_c$\;
	\While{$s$ is not a terminal state}{
		// Perform forward rollout of $\pi_h$\\
		$s' \gets T(s,\pi_h(s))$\;
		$r_c \gets r_c + R(s,\pi_h(s), s')$\;
		$s \gets s'$\;
		\lIf{$r_c \leq R_L$}{ // Reward too low\\
		\hspace{0.3cm}    $\bar{S} = \bar{S} \cup s'$; // Track state for later \\
		\hspace{0.3cm}    $R^*_h(s,\pi_h(s), s') \gets  R(s,\pi_h(s), s')$\;
		\hspace{0.3cm} $\textbf{break}$}
	}
	}
	$\pi_h^* \gets $policy trained on $R_h^*$; $\pi^* \gets $policy trained on $R$\; 	
 	Set\_Cover, Objective $\gets$ predicate\_selection$(\bar{S}, \pi^*_h, ...)$\;
 	\lIf{objective is no\_solution}{exit}
	\eIf{objective $\geq$ L (from Eq. \ref{eq:objective})}{
    \For{\text{rollout in range(1 to k)}}{
   $s \gets s_c$\;
   \While{$s$ is not terminal}{
		// Perform forward rollout of $\pi^*$\\
		$s' \gets T(s,\pi^*(s))$\;
		\lIf{$s'\in$ Set\_Cover}{$ R(s,\pi^*(s), s') \gets - L$}
		$s \gets s'$\;
	}
	}
     \textbf{go to 1}\;
   }{
   feedback $\gets$ ``There's a bad reward in \emph{Set\_Cover}.''
	
	\Return{feedback}
   }
\end{algorithm}

\subsection{Algorithm}
Here we outline the details of SPEAR for finding the predicates to communicate which will elicit optimal behavior, as presented in Algorithm  \ref{algo-reward-repair}. We continue using our running example of an emergency evacuation scenario (Figure \ref{true-map}-right) to build intuition and illustrate the inner workings of our algorithm for a human-robot collaborative scenario.

 Given an estimate of the human agent's reward function, SPEAR communicates a reward update in an attempt to elicit the best possible policy. To achieve this, (Line 3-16) we perform multiple forward rollouts of the agent policy $\pi_h$ derived from $R_h$ and (Line 13-16) compare the accumulated expected reward to the reward threshold $R_L$. The moment this threshold is crossed, the reward from that transition is determined to be relevant for updating $R_h$ and the state is added to the set cover.

This can be easily illustrated for our example, where Figure \ref{fig:policy-elicitation}b shows the belief of an agent trying to evacuate the building. Figure \ref{fig:policy-elicitation2}b gives insight into how updating the agent's reward function can result in an optimal policy even if the agent's belief about the fires doesn't truly match the environment.

After finding the states responsible for meaningful reward divergence, we find a minimal set cover of communicable predicates that map onto these states. This is achieved through \textbf{SPEAR-IP}, discussed in Section \ref{ip-formulation}. We use a third-party optimizer \cite{gurobi} to solve SPEAR-IP (Line 18), where the $predicate\_selection$ method takes in the states to cover ($\bar{S}$) and the best agent policy $\pi^*_h$. This gives a set of communicable predicates and a final objective value using Algorithm \ref{algo-predicate-selection}.

Line 19 checks whether or not a solution exists for a given set of predicates. In lines 20-28, our algorithm evaluates case 3 to determine if alternate solutions exist. In lines 23-27, SPEAR performs multiple forward rollouts of the optimal policy to find states responsible for a high objective value. In line 26, these states are penalized in the true reward function ($R$) to incentivize the algorithm to find an alternate solution which avoids these states. This enables SPEAR to explore alternate solutions, making it more robust in applications where the available predicates are insufficient, overcoming barriers due to imprecise or unavailable language. 

In line 28, now that all the appropriate states are penalized, the algorithm repeats the whole procedure from the beginning with a modified $R$, continuing this process until it finds a low objective solution or no solution (case 1 or 2). Finally, in line 30, the update is serialized as semantic feedback using the $Set\_Cover$. This feedback generation strategy uses negative reward to drive an agent's policy away from undesirable states, as improved policies can then be elicited through the exclusion of states along the agent's (hypothesized) originally intended path. While similar outcomes can be achieved via positive reward, a state exclusion-based strategy generally allows for the use of less precise predicates.

In Algorithm \ref{algo-predicate-selection}, we produce the set cover for communicating the reward update. Lines 4-7 evaluate states we want the agent to traverse (the desired trajectory $\bar{O}$) by performing a forward rollout of the best attainable agent policy $\pi^*_h$. In lines 8-13, the matrices \textit{U} and \textit{V} from Equation \ref{eq:a_matrix}-\ref{eq:b_matrix} are defined, which form the basis of the constraints governing the inclusion of states to cover and exclusion of optimal states (when giving information about negative reward) in the set cover respectively. Finally, in line 15, \emph{SPEAR-IP} is solved using the matrices \textit{U} and \textit{V} to give $Set\_Cover$ and \textit{Objective}. 

\begin{algorithm}[!ht]
	\caption{Predicate selection (Minimal set cover)}
	\label{algo-predicate-selection}
	\KwIn{\textcolor{black}{Set of States to cover $\Bar{S}$, Agent policy $\pi$, MDP  $(S,A,T)$, Set of predicates $\bar{P}$, Current state $s_c$ }}
	\KwOut{Set\_Cover and Objective (high, low, or no solution)}
	$\text{Set\_Cover} \gets \emptyset$; // Predicates in min set cover\\
	$O \gets \emptyset$; // States we don't want to cover\\
	$s \gets s_c;	O \gets O \cup s$\;
	\While{$s$ is not terminal}{
		// Perform `optimistic' forward rollout of $\pi$\\
		$s \gets \text{most likely transition from } T(s,\pi(s))$\;
		$O \gets O \cup s$; //append occupied states\\
	}
	
	//Define matrix $U$ to be $|\bar{S}| \times |\bar{P}|$ matrix s.t. \\
    \textbf{for} $i \in [1, |\bar{S}|]$, $j \in [1, |\bar{P}|]$ \\
    $u_{ij}= 
    \begin{cases}
        1,& \text{if } s_i \in \bar{S} \text{ is covered by } p_j \in \bar{P}\\
        0,              & \text{otherwise}
    \end{cases}$\\
    //Define matrix $V$ to be $|\bar{O}| \times |\bar{P}|$ matrix s.t. \\
    \textbf{for} $k \in [1, |\bar{O}|]$, $j \in [1, |\bar{P}|]$ \\
    $v_{kj}= 
    \begin{cases}
        1,& \text{if } o_k \in \bar{O} \text{ is covered by } p_j \in \bar{P}\\
        0,              & \text{otherwise}
    \end{cases}$\\

//For SPEAR-IP: refer to Equations \ref{eq:objective} - \ref{eq:constraint}\\
$\text{Set\_Cover, Objective} \gets \text{SPEAR-IP}(U,V)$;\\

	\Return{$\text{Set\_Cover, Objective}$}
\end{algorithm}

\section{Evaluation and Discussion} \label{evaluation-and-discussion}

To demonstrate the utility of our algorithm we present two applications in which an agent is required to augment the policies of other agents (autonomous or human) via policy elicitation from SPEAR. We also provide a characterization of SPEAR's performance as a function of domain size and predicate count. Results are generated on an Intel(R) Core(TM) i7-7700K CPU @ 4.20GHz.

\subsection{Robotic Cleaning Task}

\begin{figure*}[t]%
	\centering
	\includegraphics[width=\textwidth]{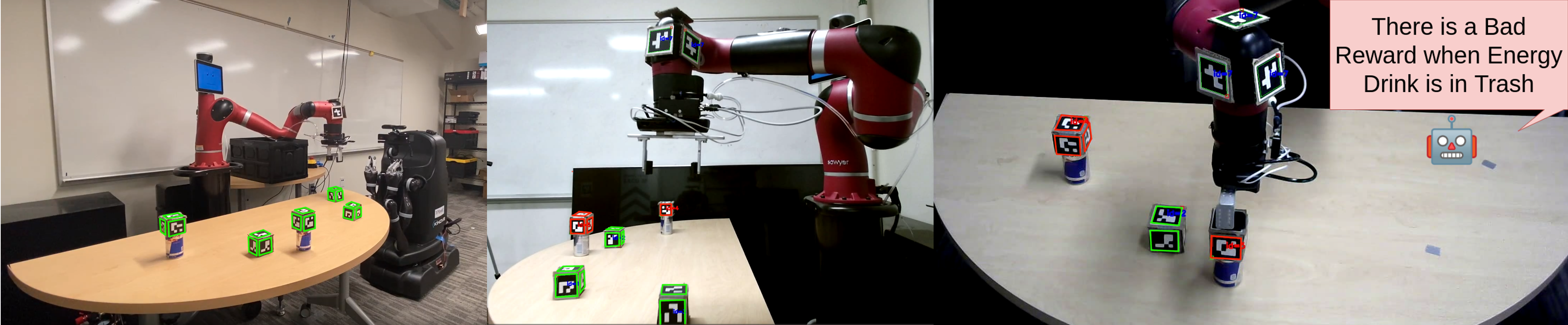}
	\caption{(Left) Sawyer (red robot) views a misinformed policy for a tabletop cleaning task as Movo (black robot) moves into position to assist. (Center) Movo, from its perspective views the scene with the correct policy and watches Sawyer remove the correct (green) objects from the table. Movo then observes Sawyer reach for one of the incorrect (red) objects and delivers a verbal reward update. (Right) Sawyer computes an update to its planner thus aligning with the desired policy. }
	\label{sawyer_movo_collaboration}
\end{figure*}

In the first scenario, one agent needs to correct the policy of a second robotic agent to prevent it from removing specific items during a pick and place task (Figure \ref{sawyer_movo_collaboration}). Importantly, these robots do not operate in the same state space, and therefore cannot directly communicate reward function updates to each other. Here, the Rethink Robotics Sawyer has been tasked with clearing \textit{trash} from a table, and is operating with the malformed belief that all objects on the table's surface are \textit{trash}. The Kinova Movo is observing this scene and has accurate knowledge about the environment (i.e., it knows that certain objects in the scene aren't trash). As Sawyer's reach expresses intent to remove one of the items that isn't trash, Movo detects that Sawyer's policy is incorrect. Movo corrects Sawyer's behavior by providing an update for its reward function, allowing Sawyer to compute a better policy. We accomplish this reward update step by generating semantic expressions about the reward function in parts of the state space (i.e., predicate-grounded natural language communicating about a region of negative reward) for Sawyer using SPEAR. 

Within this task, Sawyer used a series of ArUco markers \cite{garrido2014automatic} to track the 6-dof poses of various objects (states) on the tabletop. Using these poses, Sawyer systematically transferred all items classified as \textit{trash} to the waste bin using an interruptable pick and place action. As Movo observes Sawyer, it detects Sawyer's end effector reaching for one of the energy drinks on the table, thus indicating the intent to remove that object, and signalling to Movo that Sawyer's policy was malformed. Using SPEAR, Movo is able infer Sawyer's erroneous belief from this observed action (that a non-negative reward is associated with putting the energy drink in the trash). Movo performs a forward rollout with the inferred policy of Sawyer to predict which state transitions with \emph{negative reward} Sawyer will reach. Next, Movo computes an updated policy for Sawyer by determining which states should be assigned negative reward and which of the available predicates are needed for communicating it with Algorithm \ref{algo-predicate-selection}.
A DNF formula of predicates that covers the desired states is then computed and communicated by Movo through natural language: ``\emph{There is a Bad Reward when energy drink is in the trash}''. Finally, Sawyer uses its own set of predicates to map the communication into its own state space, update its reward function accordingly, and reconverge a repaired policy.

\subsection{Emergency Evacuation}
In our second scenario, our autonomous agent must help a human agent escape from a smart building in which a series of fires has broken out, as shown in Figure \ref{fig:policy-elicitation}. While people in the building are not aware of the fire locations, they are assumed to understand the generated predicate-based language.

The building layout for each trial is generated with a stochastic placement of rooms, hallways, and exits over a gridworld of fixed size. For example, a gridworld of size 40X40 (1600 states) may have parameters: rooms = 10, hallways = 40, number of exits = 5. These can be changed depending on the desired architecture. We then use these areas to create predicates which are grounded in the layout of the building. In our validation, we also use randomly generated predicates to prove the generalizability of our approach. 

To accommodate the full range of possible state regions to cover using predicates, we create composite predicates by creating \textcolor{black}{the power set of base predicates}. By combining base predicates into composite predicates, each scenario has an upper bound of $2^n - 1$ total predicates. 

Once the building layout and predicates are generated, we observe a randomly placed human agent explore a hazard-free building over a fixed number of episodes. The human agent's policy is then trained to always seek the shortest path to the closest exit that was discovered during these exploration episodes. Initially, SPEAR has no knowledge about which exits the human is aware of, but gradually, its belief about the human's reward function is updated from these observations. Next, we begin the evaluation by adding fire in the building via random placement and expansion. Once this process completes, we start the SPEAR evaluation. Predicates from SPEAR-IP are used to update the human agent's reward function during the episode. We update the policy of the human using the repaired reward function after each update. 

\subsection{Analysis}
We evaluated the performance of our algorithm as a function of state space size and the number of predicates using different building layouts. We also demonstrated that our algorithm can generalize to arbitrary state mappings by using randomly generated predicates. To fairly evaluate this, we randomly generate ball-shaped predicates (evaluating to true within a given radius of a random state vector). When in a structured environment, the predicates grounded in the layout of the building tend to naturally make it easier to find potentially hazardous predicates. For example, in Figure \ref{fig:policy-elicitation2}a, predicates are grounded in the layout of the building and fire can be avoided through communicating the \textit{``Center Hallway''} predicate as it is directly associated with pathway of the human. However, if predicates are placed stochastically throughout the map (corresponding to cases where the language isn't tailored to the domain), finding a solution becomes more difficult. Hence, by using randomly placed predicates we make it difficult for the algorithm to find a low cost solution while also directly increasing the computational time, providing performance analysis for cases of predicate-domain mismatch. 

\begin{figure}[t]%
	\centering%
	\includegraphics[width=0.6\textwidth]{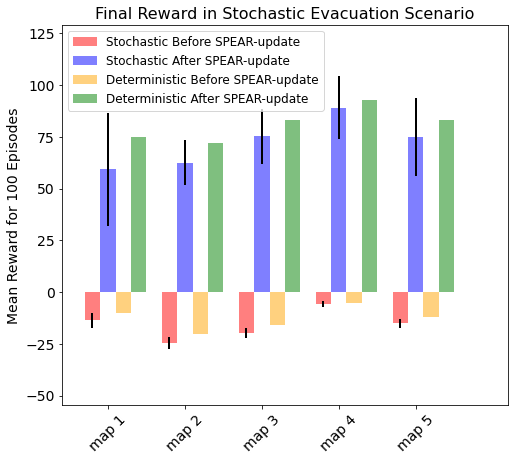}
	\caption{The evaluation of SPEAR algorithm in stochastic and deterministic evacuation domains (25x25). Substantial increases in episode reward occur due to the generated symbolic reward updates.}
	\label{stochastic-plot}
\end{figure}

We qualitatively assess our algorithm in an emergency evacuation scenario for both deterministic and stochastic environments (Figure \ref{stochastic-plot}). In the stochastic domain, a stochastic transition function was applied describing the model's action-based state transition dynamics. For each map (25 X 25) in both environments, we evaluated our algorithm over the course of 100 episodes. In each episode the reward was computed both before and after the SPEAR update (exit: +100, fire: -100, and each step: -1 ). At each time step, our agent would take the prescribed action with a probability of 85\% and would alternatively take a step in a different random direction. For our stochastic evaluation, we set our forward rollout count parameter, $k$, to 10. The results from our simulations show a substantial improvement in the episodic reward after the SPEAR update, (Figure \ref{stochastic-plot}) demonstrating its utility in both deterministic and stochastic domains.

Furthermore, we do performance analysis as a function of predicate count by dividing into two success cases described in Section \ref{ip-formulation}: 1) Maps with a low objective value solution (Case 1); and 2) Maps with high objective value solution (Case 3). 

\begin{figure*}[t]
	\centering
	\includegraphics[width=\textwidth]{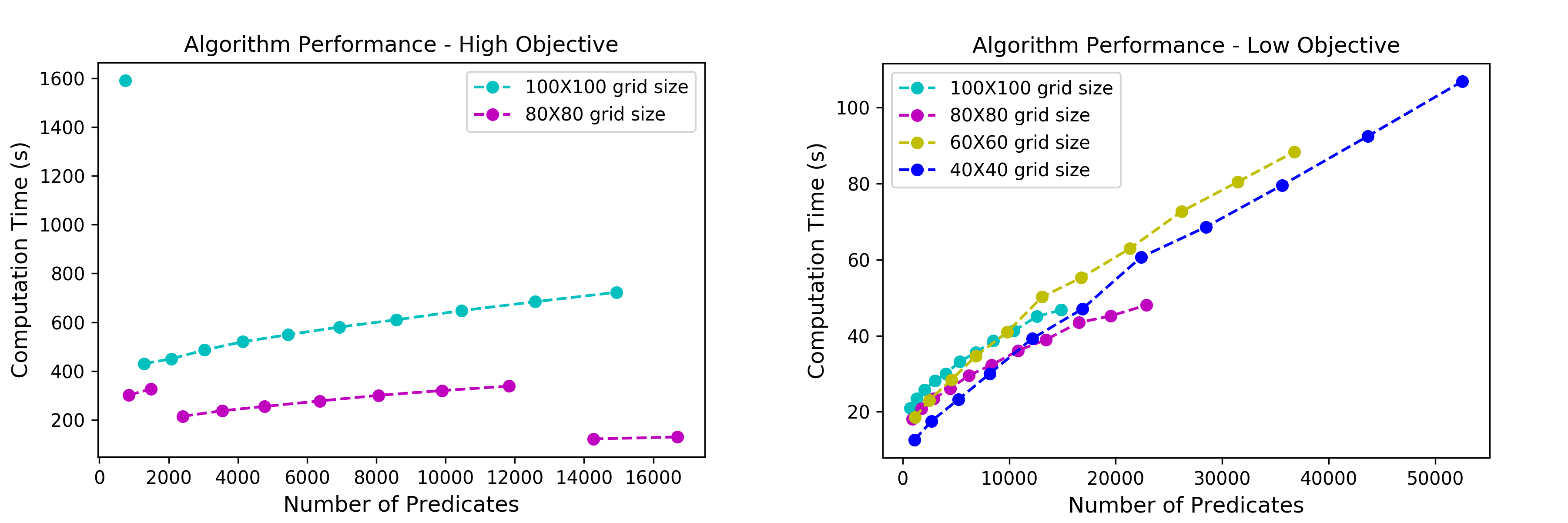}
	\caption{(Left) SPEAR's performance for high-objective (case 3) maps scales linearly as predicate count increases. Sharp decreases in computation time (purple) occur when new predicates cause SPEAR to find solutions in fewer algorithm loops. (Right) SPEAR's performance for low objective (case-1) maps. 
Plots reveal a linear relationship between runtime and predicate count.}
	\label{spear-performance-predicates}
\end{figure*}

For each map of a given size, we start with 100 base predicates and increase in increments of 10. Figure \ref{spear-performance-predicates}-right shows how algorithm performance changes with increasing predicate count on low objective value maps. For Case 1 solutions our algorithm exhibits linear computational costs as a function of the number of predicates, and the set cover can be computed within 50 seconds with nearly 10,000 communicable predicates to choose from. To provide context for the significance of this result, we can consider a comparison against the state-of-the-art method for succinctly describing state regions (using the Quine-McCluskey algorithm) by Hayes and Shah \cite{hayes2017hri}.  Our Integer Programming approach to this set cover problem achieves a several order of magnitude improvement over the QM based method (where set covers with 10 possible predicates takes ~60-120 seconds to compute on similar hardware, deteriorating exponentially as a function of predicate count). SPEAR is made possible by this performance improvement, effectively operationalizing the insight of Hayes and Shah's work as a means of communicating information about those state regions --- enabling SPEAR's policy update method.

Similarly, we plot the performance for Case 3 solutions as predicate count increases (Figure \ref{spear-performance-predicates}-left). We observe that the plots again scale linearly in practice with respect to the number of predicates, but with higher computation time due to multiple SPEAR runs. Computation time is higher here because the algorithm has to explore alternative solutions for the desired policy, solving for set cover solutions multiple times (Section \ref{ip-formulation}).

Our method achieves a dramatic improvement in computation time over the QM based set cover approach for two reasons: offline computation of predicates and allowing approximate solutions. QM uses a two-step procedure: 1) finding prime implicants, and 2) using those prime implicants to find the exact minimal set cover (see \cite{mccluskey1956minimization}). In SPEAR, we alleviate a significant time sink by pre-computing predicate values over the state space (which the QM approach had to solve online in step 1). Secondly, since SPEAR is often able to use approximate set covers to communicate its reward updates (i.e., sometimes it's okay to over-state or under-state the reward region, as in Figure \ref{fig:policy-elicitation2}b), we are able to relax the requirement of having an exact set cover. In contrast, QM solves for exact set cover making it slower in execution and limited in application. For example, referring to Figures \ref{fig:policy-elicitation}c and \ref{fig:policy-elicitation2}a, QM will try to find the precise set cover for covering only the two fire states in the central hallway (and fail as there is no exact solution). Whereas, SPEAR will try to find the approximate set cover that encompasses those fire states, accepting inaccuracy if necessary so long as it still achieves the same outcome as if there were a perfect solution (Case 1) and preferring a suboptimal solution to no solution (Case 3). This added layer of approximation heuristic makes SPEAR faster and more flexible than the QM based approach.

Interestingly, we found some irregularities in performance as illustrated within the 80 X 80 world (Figure \ref{spear-performance-predicates}-left-purple). We found that a sudden dip in computation time can occur when new language enables the algorithm to find a Case 1 (single-loop of algorithm) or cheaper Case 3 (still multiple loops of the algorithm, but fewer) solution. 

\begin{figure*}[t]
	\centering
	\includegraphics[width=0.6\textwidth]{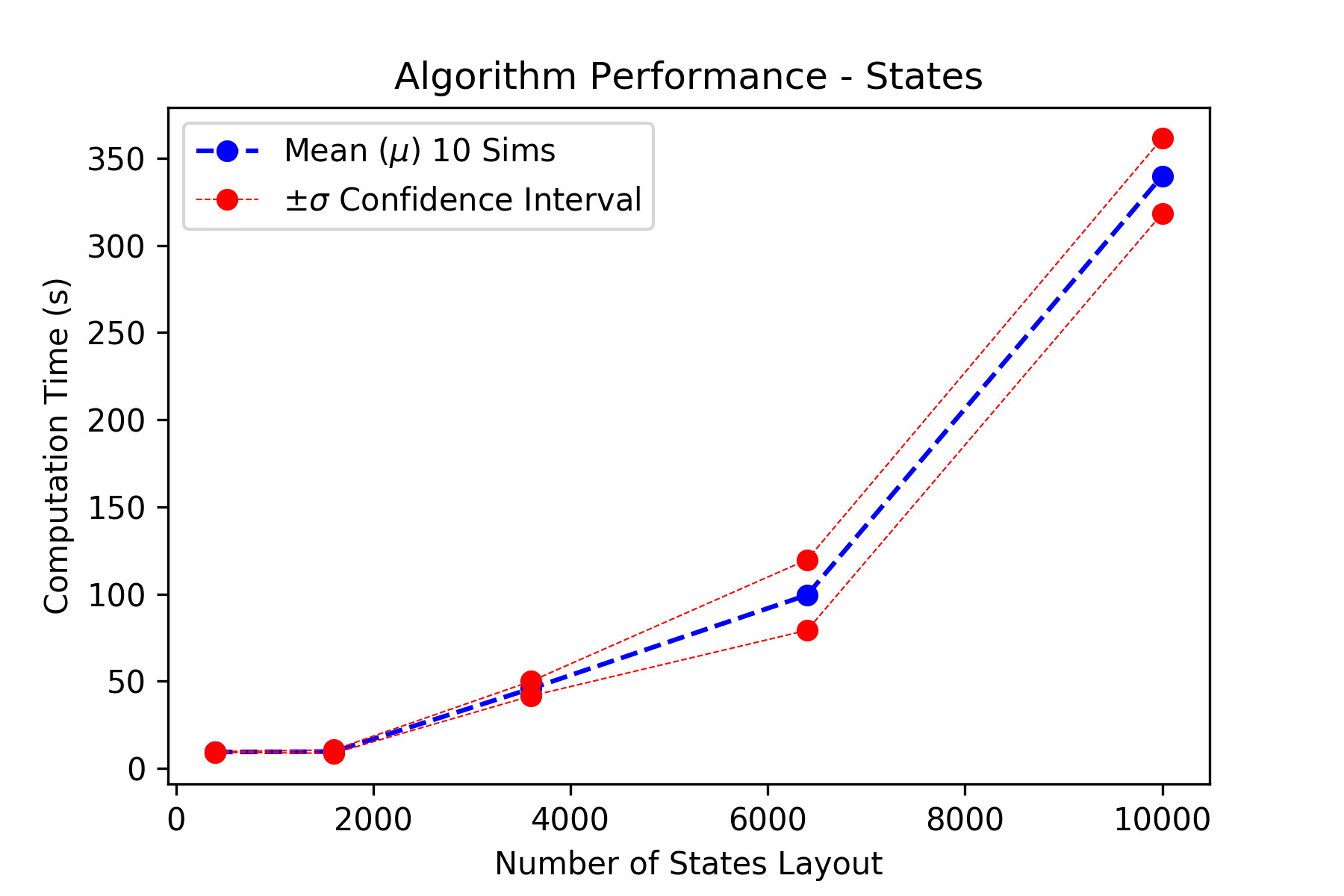}
	\caption{ The evaluation of the SPEAR algorithm in stochastic and deterministic evacuation domains (25x25). SPEAR's performance as state space size increases using 100 randomly generated base predicates. Computation time increases polynomially as a function of state space size}
	\label{spear-performance-states}
\end{figure*}

The final part of our analysis looks at the performance of our algorithm as domain size increases with fixed predicate count (100 base predicates). We generate a set of maps with similar parameters for a fixed state space size, sampling from these maps to get the mean and confidence bound for 10 simulation runs as shown in Figure \ref{spear-performance-states}. A significant take-away from this analysis is the insight that an attention mechanism is more important for abstracting and reducing the domain's state space than for limiting the number of predicates to consider, as prior work anticipated \cite{abel2018state,hayes2017hri,tabrez2019explanation}.

We have shown that \emph{SPEAR} enables dramatic improvements in agent performance through policy elicitation, while also contributing a novel method for communicating about state regions that substantially outperforms prior work. These contributions enable semantically guided policy manipulation for a much broader class of problems than was previously possible, providing a method that scales linearly with predicate count as opposed to exponentially, advancing the state-of-the-art in autonomous coaching through new algorithms and improved foundational capability.

\section{Conclusion} \label{conclusion}

In this work, we describe an approach for \emph{policy elicitation}, the manipulation of an agent's behavior through the use of semantically grounded reward updates. We present a novel Integer Programming-based algorithm for rendering policy explanation \cite{hayes2017hri} and policy manipulation \cite{tabrez2019explanation} techniques feasible for use in applications substantially larger than previously possible. We demonstrate a several order of magnitude improvement over the state-of-the-art in terms of runtime and allowable domain size when communicating about state regions, and utilize this to achieve significant improvements in agent performance through policy elicitation from an autonomous coaching agent.

\bibliographystyle{spmpsci}
\bibliography{main}
\end{document}